\documentclass[letterpaper, 10 pt, conference]{ieeeconf}  
\IEEEoverridecommandlockouts    
\usepackage{comment}
\usepackage{cite}
\newcommand{\eat}[1]{}
\usepackage{booktabs}
\usepackage{color}
\usepackage [autostyle, english = american]{csquotes}
\MakeOuterQuote{"}
\usepackage{xcolor}
\let\labelindent\relax 
\usepackage{enumitem}

\usepackage{lipsum}  
\usepackage{blindtext}

\usepackage{multicol}
\usepackage{multirow}
\usepackage{diagbox}    
\usepackage{mathtools}
    
\usepackage{amsmath}
\usepackage{amstext}
\usepackage{amssymb, wasysym}
\usepackage{amsfonts}
\usepackage{float}

\usepackage{amsthm}  
\usepackage[normalem]{ulem} 

\newtheorem{remark}{Remark}

\newtheorem{assumption}{Assumption}
\newtheorem{problem}{Problem}
\usepackage{subcaption}
\usepackage{caption}
\usepackage{todonotes}

\usepackage[extramath,probability,reviewmark]{mylatexdefs}

\makeatletter

\newcommand{\Rmnum}[1]{\expandafter\@slowromancap\romannumeral #1@}
\usepackage{savesym}

\usepackage{algorithm}
\usepackage{algorithmicx} 
\usepackage{algpseudocode} %
\savesymbol{AND}
\usepackage[group-separator={,},group-minimum-digits={3}]{siunitx}

\usepackage{graphicx} 
\usepackage{epsfig} 

\usepackage{times} 
\usepackage{amsmath} 
\usepackage{amssymb}  
\usepackage{comment}
\makeatletter
\let\NAT@parse\undefined
\makeatother
\usepackage{hyperref}
\hypersetup{
   colorlinks=true,
    linkcolor= blue,
    allcolors=blue,
    citecolor = blue,
    filecolor=black,      
    urlcolor=blue,
    }
\usepackage{mathrsfs}

\usepackage[symbol]{footmisc}

\newcolumntype{K}[1]{>{\centering\arraybackslash}p{#1}}

\title{\LARGE \bf
Social Navigation in Crowded Environments with Model Predictive Control and
Deep Learning-Based Human Trajectory Prediction}
\author{Viet-Anh Le$^{1,2}$, Behdad Chalaki$^{3*}$, Vaishnav Tadiparthi$^{3*}$,\\ Hossein Nourkhiz Mahjoub$^3$, Jovin D'sa$^3$, and Ehsan Moradi-Pari$^3$
\thanks{\scriptsize$^*$Both authors contributed equally.}
\thanks{\scriptsize$^1$Department of Mechanical Engineering, University of Delaware, DE 19716 USA (Email: vietale@udel.edu).
\scriptsize$^2$System Engineering, Cornell University, NY 14850 USA.
This work was conducted during V.-A. Le's internship at Honda Research Institute.}
\thanks{\scriptsize$^3$Honda Research Institute USA, Inc. (Email: \{behdad\_chalaki; vaishnav\_tadiparthi;  hossein\_nourkhizmahjoub; jovin\_dsa; emoradipari \}@honda-ri.com).}
}
\begin{document}

\maketitle
\thispagestyle{empty}
\pagestyle{empty}

\begin{abstract} 
Crowd navigation has received increasing attention from researchers over the last few decades, resulting in the emergence of numerous approaches aimed at addressing this problem to date. 
Our proposed approach couples agent motion prediction and planning to avoid the freezing robot problem while simultaneously capturing multi-agent social interactions by utilizing a state-of-the-art trajectory prediction model i.e., social long short-term memory model (Social-LSTM). Leveraging the output of Social-LSTM for the prediction of future trajectories of pedestrians at each time-step given the robot's possible actions, our framework computes the optimal control action using Model Predictive Control (MPC) for the robot to navigate among pedestrians.
We demonstrate the effectiveness of our proposed approach in multiple scenarios of simulated crowd navigation and compare it against several state-of-the-art reinforcement learning-based methods.
\end{abstract}
\section{Introduction}

\PARstart{I}{n} various disciplines of robotics, encompassing autonomous driving, manipulation, navigation, \textit{etc.}, the most challenging scenarios arise when robots are required to to co-exist with humans. 
This is mainly due to stochasticity in humans' behavior which makes it very difficult for the robots to confidently plan and execute their tasks.

In the realm of robot navigation, the problem we tackle in this paper is typically referred to as \textit{social navigation}, a.k.a. \textit{crowd navigation}. 
It has received a great deal of attention during the last decade \cite{KRUSE20131726,CHARALAMPOUS201785,francis2023principles,mavrogiannis2023core} and aims to enable robots to achieve their navigation goals via interacting
with their surrounding agents, \textit{i.e.,} humans or other robots, in a way that those surrounding agents do not go through an unpleasant experience during these interactions. 
This is a complex problem due to the several factors that are contributing to the quality of these inter-agent interactions, among which safety, comfort,
legibility, politeness, social competency, agent understanding,
pro-activity, and responsiveness to context have been considered as the most important parameters by the social navigation research community \cite{francis2023principles}.

Different solutions have been proposed in the literature to address some of these requirements. However, one can divide the overall current literature on social navigation into three main categories of solutions: 1) Reactive \cite{van2008reciprocal}; 2) Reinforcement Learning (RL); and 3) Optimization.

Reactive-based methods in general are designed based on considering other agents as moving obstacles and taking into account their reactive behaviors with some assumptions on their collision avoidance strategies. 
On the other hand, several studies have utilized trajectory-based RL frameworks with deep neural networks to address the aforementioned requirements. 
Some of the most important RL-based methods are Collision Avoidance with Deep RL (CADRL) \cite{chen2017socially}, LSTM-RL \cite{everett2018motion} which is aimed at handling arbitrary numbers of agents, and SARL \cite{chen2019crowd} for obtaining the collective impact of crowd through a self-attention mechanism, and recurrent graph neural network with attention mechanisms \cite{liu2023intention}. 
All of these efforts seek to train navigation policies for a single robot that maximizes a specially designed reward function while minimizing the possibility of collisions with other agents. 
However, the main challenge of RL-based methods is their vulnerability to cases that haven't been encountered during the policy's training phase.
The third category of methods, \textit{i.e.,} optimization-based frameworks such as Model Predictive Control (MPC) can be typically used to optimize the behavior of a robot over a finite control horizon assuming there are reliable prediction models for human trajectories in hand.
For instance, Brito \textit{et.al.,} \cite{brito2021go} combined RL with an optimization-based method in which a learned policy provides long-term guidance to a local MPC planner. 
Several other studies have recently proposed MPC with various human prediction models, including constant velocity \cite{akhtyamov2023social}, intention-enhanced optimal reciprocal collision avoidance (iORCA) \cite{chen2021interactive}, 
social generative adversarial networks (GAN) \cite{poddar2023crowd},  long short term memory (LSTM) \cite{lindemann2022safe}, and Kalman filters
\cite{vulcano2022safe}.
Recent advances in machine learning-based human trajectory prediction have opened new doors for social navigation problems and several of these have demonstrated the superiority of their prediction techniques over the previous works. 
Some of the most important ML-based prediction frameworks which have shown their potential in the context of social navigation are Social-LSTM \cite{alahi2016social}, Social-GAN \cite{gupta2018social}, Social-NCE \cite{liu2021social}, and sparse Gaussian processes \cite{trautman2017sparse}.
Therefore, due to the superiority of these machine learning prediction models, their combination with MPC framework could, in theory, improve planning performance.

In this paper, we present our framework for robot navigation in crowded environments in which we integrate a machine-learning based trajectory prediction model \textit{i.e.,} Social-LSTM \cite{alahi2016social} into an optimization-based planning in an MPC fashion. 
We couple the prediction and planning to avoid freezing robot problem \cite{trautman2015robot} while capturing multi-agent social interactions among the robot and pedestrians. 
In our framework, we leverage a Social-LSTM model trained on a real human-trajectory data-set to predict the future behavior of human pedestrians and their interactions with the robot's possible actions. 
The solution of MPC framework coupled with the Social-LSTM model is the optimal control action for the robot to navigate among the crowd.
To numerically solve the MPC problem coupled with the Social-LSTM , we utilize an iterative best-response (IBR) approach \cite{espinoza2022deep} inspired by the Nash equilibrium \cite{bacsar1998dynamic}. 
At each time-step, the method sequentially computes the neural network prediction and solves for the optimal control action of the robot. 
The performance of the proposed method is evaluated in simulations with different scenarios in comparison with baseline RL techniques to demonstrate the potency and the domain-invariant nature of the MPC approach.

This paper makes two main contributions to the body of literature. Our first contribution involves addressing the freezing robot problem by incorporating the Social-LSTM prediction model into the MPC framework in a recursive fashion, hence enabling a coupled prediction and planning. 
Second, to the best of our knowledge, in the context of crowd navigation, this study is the first that considers leveraging the iterative best response approach for solving the problem. 

The remainder of the paper is organized as follows.
In Section~\ref{sec:problem}, we present our problem statement for crowd navigation, while we provide the specific details of the proposed framework in Section~\ref{sec:mpc}.
We present our simulation results in multiple benchmark scenarios together with analysis in Section~\ref{sec:sim}. Finally, we draw concluding remarks and propose some directions for future research in Section~\ref{sec:conclu}.

\section{Problem Statement}
\label{sec:problem}

\begin{figure}
\centering
\includegraphics[width=0.75\linewidth, bb = 250 140 550 480, clip=true]{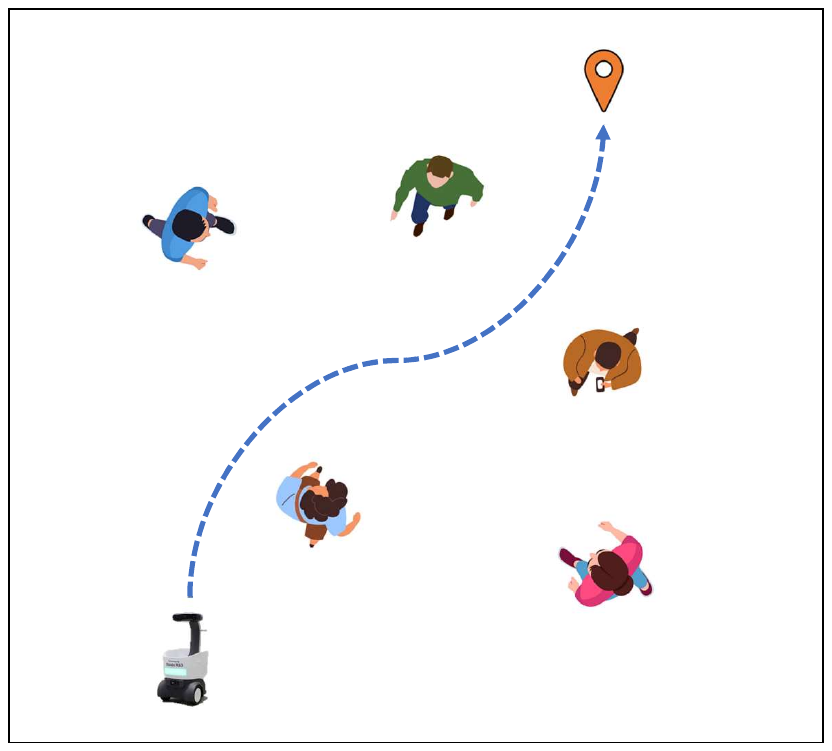}
\caption{An example of a robot navigating in a crowded environment.}
\label{fig:example}
\vspace{-7mm}
\end{figure}

We consider an environment $\WWW \subset \RR^2$ where a single robot navigates among $N\in\NN$ human pedestrians, as can be illustrated in Fig.~\ref{fig:example}.
Let $0$ be the index of the robot, while $\HHH = \{ 1, \dots, N \}$ denotes the set of human pedestrians in the environment.  

At time-step $k\in\NN$, let $\bb{s}_{0,k} = [s^x_{0,k}, s^y_{0,k}]^\top \in \WWW$, $\bb{v}_{0,k} = [v^x_{0,k}, v^y_{0,k}]^\top \in \RR^2$, and $\bb{a}_{0,k} = [a^x_{0,k}, a^y_{0,k}]^\top \in \RR^2$ be the vectors corresponding to position, velocity, and acceleration of the robot in Cartesian coordinates, respectively, where each vector consists of two components for $x-$ and $y-$ axis. 
Additionally, the robot needs to navigate from an initial position $\bb{s}^\mathrm{orig}_{0}\coloneqq [s^x_{0,0}, s^y_{0,0}]^\top$ called origin to a final goal position $\bb{s}_{0}^\mathrm{goal} \in\mathcal{W}$ while avoiding collisions and any potential discomfort to other pedestrians $i\in\mathcal{H}$. 
Discomfort is a social conformity metric and is defined to be present if the robot's projected path intersects with a human's predicted path \cite{wang2022metrics}.
Let $\bb{x}_{0,k}^\top = [\bb{s}_{0,k}^\top, \bb{v}_{0,k}^\top]$ and $\bb{u}_{0,k} = \bb{a}_{0,k}$ be the state vector and control action for the robot at time-step $k$, respectively.  
Likewise, let $\bb{s}_{i,k} =[s^x_{i,k}, s^y_{i,k}]^\top\in \WWW$ be the position of human $i\in\mathcal{H}$ at time-step $k$ in a vector form.
\begin{assumption}
We assume that the real-time position of each pedestrian can be determined through the use of onboard sensors or by obtaining data from a positioning system.
\end{assumption}


\section{Robot Navigation with Model Predictive Control}
\label{sec:mpc}

Our framework consists of two main components: (1) a Social-LSTM model \cite{alahi2016social}  which learns the social interaction and predicts the future behavior of human pedestrians, and (2) an MPC to find the optimal control action for the robot.

\subsection{Human Motion Prediction using Social-LSTM}

Let $t \in \NN$ be the current time-step, $H \in \mathbb{Z}^+$ is the control/prediction horizon length (with both equal to each other), and $\III_t = \{t,t+1,\dots,t+H-1\}$ be the set of time-steps in the control horizon.
The human prediction model aims at predicting the trajectories of human pedestrians over a prediction horizon of length $H$ given the current and past observations over $L \in \mathbb{Z}^+$ previous time-steps of all agents' trajectories including the robot's.
Social-LSTM was developed in \cite{alahi2016social} for jointly predicting multi-step trajectory of multiple agents.
It uses a separate LSTM network for each trajectory, then the LSTMs are connected to each other through a social pooling (S-pooling) layer.

We consider recursive prediction for the pedestrians' positions over the next control horizon using the single-step Social-LSTM model denoted by {$\phi(\cdot):\mathbb{R}^{2(N+1)(L)}\rightarrow\mathbb{R}^{2N}$} as follows:
\cite{gupta2023interaction}
\begin{equation}
\label{eq:lstm_recursive}
\bb{s}_{1:N,k+1} = \phi (\bb{s}_{0:N,k-L+1:k}),\, \forall k \in \III_t.
\end{equation}

In \eqref{eq:lstm_recursive}, at each time-step, predicted positions of pedestrians computed from the previous time-steps are used recursively as the inputs of the Social-LSTM model. 
Furthermore, the Social-LSTM-based predicted positions of the robot are disregarded as they are computed using the solution of the MPC problem.
For further details on the architecture design and implementation of Social-LSTM, the readers are referred to \cite{alahi2016social}.
\begin{remark}
    It should be noted that while in this work we employ the Social-LSTM model \cite{alahi2016social} as a human prediction model, our framework can be integrated with alternative deep learning models such as \cite{gupta2018social}, \cite{liu2021social}, \cite{kothari2021human}.
\end{remark}


\subsection{Model Predictive Control for Crowd Navigation}

In this section, we formulate an MPC problem to navigate the robots while taking into account the trajectory prediction model of surrounding pedestrians.
For ease of notation, henceforth, we use $\bb{u}_0$, $\bb{x}_0$, and $\bb{s}_i$, $\forall i \in \HHH$ instead of $\bb{u}_{0,t:t+H-1}$, $\bb{x}_{0,t+1:t+H}$ and $\bb{s}_{i,t+1:t+H}$, respectively, to denote the vectors concatenating the variables over the control horizon.

The system dynamics of the robot  for all $k \in \III_t$ is given by the following discrete-time double-integrator model
\begin{equation}
\label{eq:dynamics}
\begin{split}
\bb{s}_{0,k+1} &= \bb{s}_{0,k} + \tau \bb{v}_{0,k} + \frac{1}{2} \tau^2 \bb{a}_{0,k}, \\
\bb{v}_{0,k+1} &= \bb{v}_{0,k} + \tau \bb{a}_{0,k},  
\end{split}
\end{equation}
where $\tau \in \RRplus$ is the sampling time period. 

The speed and control input of the robot at each time-step $k$ are bounded by:
\begin{equation}
\label{eq:bound}
\begin{split}
-v_{\max} & \le v^x_{0,k}, v^y_{0,k} \le v_{\max},  \\    
-a_{\max} & \le a^x_{0,k}, a^y_{0,k} \le a_{\max},
\end{split}
\end{equation}
where $v_{\max} \in \RRplus$ and $a_{\max} \in \RRplus$ are the maximum velocity and maximum acceleration, respectively.
We formulate the total objective function in MPC by a weighted sum of multiple distinct objectives, representing a diverse set of performance criteria for the robot.
In particular, to navigate the robot to the goal, we include tracking minimization to the desired trajectory 
\begin{equation}
J^\mathrm{goal} (\bb{s}_0) = \sum_{k=t}^{t+H-1} (\bb{s}_{0,k+1} - \bb{s}_{0,k+1}^{\mathrm{ref}})^\top (\bb{s}_{0,k+1} - \bb{s}_{0,k+1}^{\mathrm{ref}}),
\end{equation}
where $\bb{s}_{0,k+1}^{\mathrm{ref}}$ is the desired position at time $k+1$.
We compute the desired trajectory based on the straight line to the robot’s goal as follows 
\begin{equation}
\! \bb{s}_{0,k+1}^{\mathrm{ref}} \! = \! \bb{s}_{0,k}^{\mathrm{ref}} 
+ \min \! \Big \{ \! \tau v_{\max}, \norm{\bb{s}_0^\mathrm{goal} - \bb{s}^{\mathrm{ref}}_{0,k}} \! \Big \}  
\frac{\bb{s}_0^\mathrm{goal} - \bb{s}^{\mathrm{ref}}_{0,t}}{\norm{\bb{s}_0^\mathrm{goal} - \bb{s}^{\mathrm{ref}}_{0,t}}}, 
\end{equation}
for $k\in\III_t$ and $\bb{s}^{\mathrm{ref}}_{0,t} = \bb{s}_{0,t}$. 

In addition, we minimize the acceleration and jerk rates of the robot's motion by the following objectives
\begin{equation}
J^\mathrm{acce} (\bb{u}_0) = \sum_{k=t}^{t+H-1} \bb{u}_{0,k}^\top \bb{u}_{0,k},
\end{equation}
and
\begin{equation}
J^\mathrm{jerk} (\bb{u}_0) = \sum_{k=t}^{t+H-1} ( \bb{u}_{0,k} - \bb{u}_{0,k-1} )^\top ( \bb{u}_{0,k} - \bb{u}_{0,k-1} ).  
\end{equation}
To encourage safety between the robot and the pedestrians, we impose the following  constraint that the distance between the robot and each pedestrian $i\in\mathcal{H}$ be greater than a safe speed-dependent distance
\begin{equation}
\norm{\bb{s}_{0,k+1} - \bb{s}_{i,k+1}}_2^2 \geq d_{\min}^2 + \rho \norm{\bb{v}_{0,k+1}}_2^2, 
\end{equation}
where $d_{\min} \in \RRplus$ is the minimum allowed distance and $\rho \in \RRplus$ is a scaling factor.
The above constraint implies that the robot should keep further distances from the humans while moving at higher speed.
We include the collision avoidance constraint as a soft constraint in the objective function by using a smoothed max penalty function as follows
\begin{equation}
\begin{multlined}
J^\mathrm{coll} (\bb{x}_{0}, \bb{s}_{i}) 
= \sum_{k=t}^{t+H-1} \mathrm{smax} \Big( d_{\min}^2 + \rho \norm{\bb{v}_{0,k+1}}_2^2 - \\ 
\norm{\bb{s}_{0,k+1} - \bb{s}_{i,k+1}}_2^2 \Big),
\end{multlined}
\end{equation}
where the smoothed max penalty function is defined as
\[  \mathrm{smax}(x) = \frac{1}{\mu} \log \big( \exp(\mu x) + 1 \big), \]
with $\mu \in \RRplus$ as a parameter that manipulates the smoothness of the penalty function.

The MPC objective function can be given by a weighted sum of those features as follows
\begin{equation}
\begin{multlined} \label{eq:TOTALCOST}
\! J (\bb{u}_0, \bb{x}_0, \bb{s}_{1:N})  
= \omega^\mathrm{goal} J^\mathrm{goal} (\bb{x}_0) + \omega^\mathrm{acce} J^\mathrm{acce} (\bb{u}_0) \\ 
\!\! + \omega^\mathrm{jerk} J^\mathrm{jerk} (\bb{u}_0) \! + \!
\sum_{i \in \HHH} \omega^\mathrm{coll} J^\mathrm{coll} (\bb{x}_0, \bb{s}_i),
\end{multlined}
\end{equation}
where $\omega^\mathrm{goal}, \omega^\mathrm{acce}, \omega^\mathrm{jerk},$ and $\omega^\mathrm{coll} \in \RRplus$ are positive weights.
Note that the penalty weight $\omega^\mathrm{coll}$ chosen should be sufficiently large.
Hence, the MPC formulation for each time-step $t$ is formulated as follows
\begin{problem} \label{pr:Problem1} 
{At time-step $t\in\mathbb{N}$, robot $0$ solves the following MPC problem, the solution of which provides the best control actions for the subsequent $H$ steps, represented as $\bb{u}_0=\bb{u}_{0,t:t+H-1}$. However, at time-step $t$, the robot alone executes the initial control action $\mathbf{u}_{0,t}$ and disregards the subsequent actions.}  
\begin{subequations}
  \label{eq:MPC}
  \begin{align}
    &
    \begin{multlined}
    \underset{ \bb{u}_0}{\minimize} \; J (\bb{u}_0, \bb{x}_0, \bb{s}_{1:N}),
    \end{multlined}
    \label{eq:MPC:obj}\\
    & \emph{subject to: } 
    \eqref{eq:lstm_recursive},
    \eqref{eq:dynamics}, \emph{and }  \eqref{eq:bound}, \, \forall k\in\III_t, \\
    & \emph{given:} 
    \quad 
    \bb{s}_{0:N,t-L+1:t}.
  \end{align}
\end{subequations}
\end{problem}

\subsection{Iterative Best-Response Implementation}

In order to solve the MPC problem \eqref{eq:MPC} coupled with the Social-LSTM model, it is possible to employ gradient-based techniques that require the computation of gradients through back-propagating the LSTM’s gradients \cite{gupta2023interaction}.
However, due to the complexity of the neural network model, solving the MPC problem would be computationally intractable.
Hence, in this section, we introduce an iterative best-response technique \cite{williams2018best,espinoza2022deep} inspired by the Nash equilibrium concept. This approach involves successively computing the neural network prediction and solving the MPC problem at each time-step for several iterations or until convergence is achieved. We use superscript $j\in\NN$ in $\bb{u}_{0}^{(j)}$ and $\bb{x}_{0}^{(j)}$ to represent the outcomes at the $j$'th iteration.
If the algorithm converges, the resultant state is considered a Nash equilibrium \cite{williams2018best,espinoza2022deep}.
The iterative best-response algorithm for solving MPC problem with the recursive prediction model is detailed in Algorithm~\ref{alg:iterative}.
At $t = 0$, we initialize $\bb{u}_{0}^{(0)} = \bb{0}$, and at every time-step $t > 1$, the optimization is warm-started with the solution of the previous time-step.

\begin{algorithm}[ht!]
  \caption{Iterative Best-Response MPC Implementation}
  \label{alg:iterative}
  \begin{algorithmic}[1]
    \Require $t$, $H$, $j_{\mathrm{max}} \in \NN$, $\epsilon \in \RRplus$, $\bb{u}_{0}^{(0)}:= \bb{u}_{0, t:t+H-1}^{(0)}$, $\bb{s}_{0}^{(0)} := \bb{s}_{0, t+1:t+H}^{(0)}$, $\bb{s}_{1:N,t-L:t}^{(0)}$  
    \For{$j = 1,2,\dots,j_{\mathrm{max}}$}
    \State Predict  $\bb{s}_{1:N}^{(j)} := \bb{s}_{1:N, t+1:t+H}^{(j)}$ recursively by  \eqref{eq:lstm_recursive} given  $\bb{s}_{0}^{(j-1)}$.
    \State Solve \eqref{eq:MPC} 
    given $\bb{s}_{1:N}^{(j)}$ to obtain $\bb{u}_{0}^{(j)}$ and $\bb{x}_{0}^{(j)}$.
    \If {$\norm{\bb{u}_{0}^{(j)} - \bb{u}_{0}^{(j-1)}} \le \epsilon$}
    \State \Return $\bb{u}_{0}^{(j)}$
    \EndIf
    \EndFor
    \State \Return $\bb{u}_{0}^{(j_{\mathrm{max}})}$
  \end{algorithmic}
\end{algorithm} 
\setlength{\textfloatsep}{0.05cm}

\section{Simulation Results}
\label{sec:sim}


\begin{figure}[h!] 
\centering
\begin{subfigure}{.235\textwidth} 
\centering
\includegraphics[width=\textwidth, bb = 10 10 560 450, clip=true]{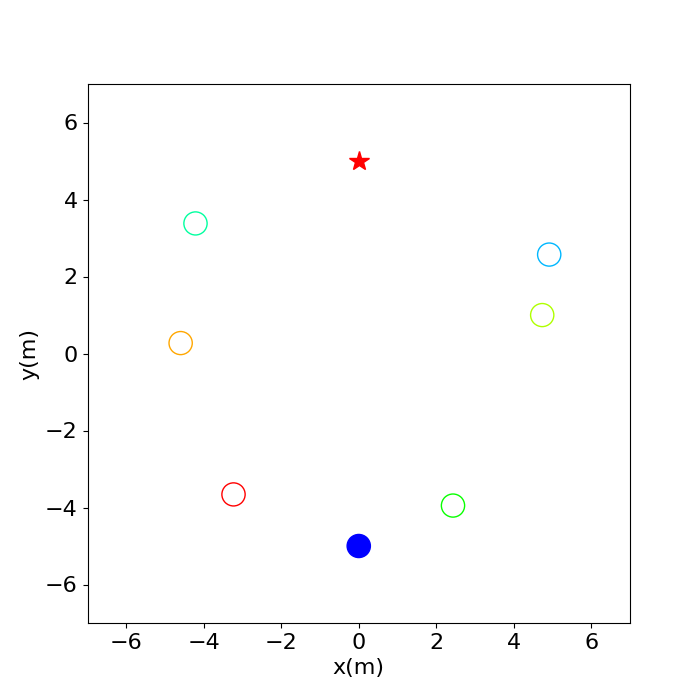}
\caption{$t = \SI{0}{s}$}
\end{subfigure} 
\begin{subfigure}{.23\textwidth} 
\centering
\includegraphics[width=\textwidth, bb = 10 10 560 450, clip=true]{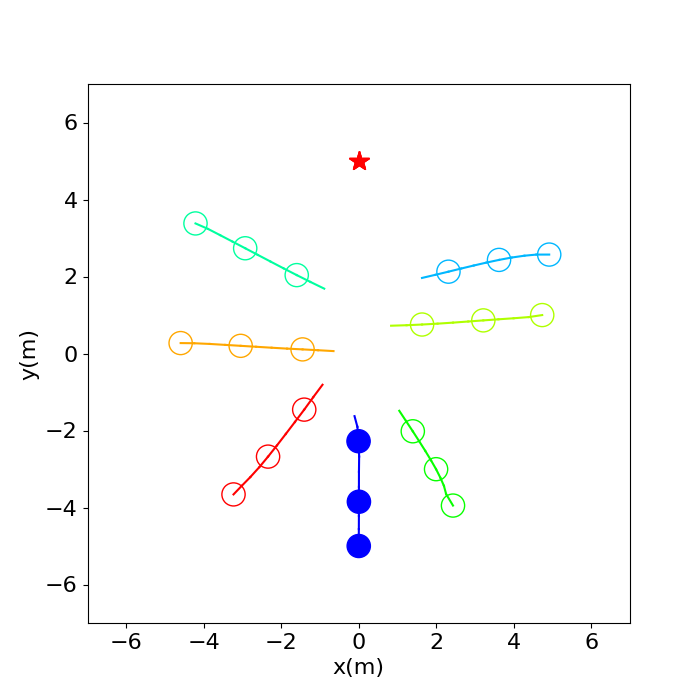}
\caption{$t = \SI{4}{s}$}
\end{subfigure} 

\begin{subfigure}{.235\textwidth} 
\centering
\includegraphics[width=\textwidth, bb = 10 10 560 450, clip=true]{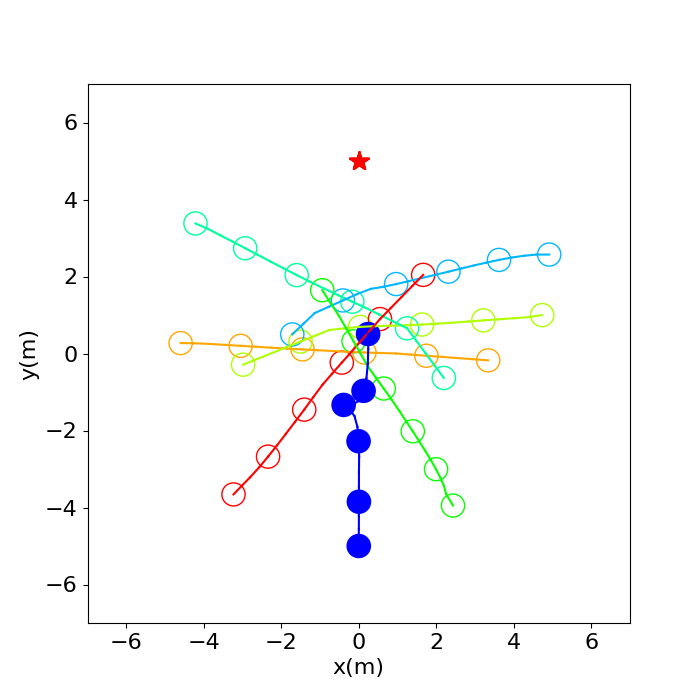}
\caption{$t = \SI{8}{s}$}
\end{subfigure} 
\begin{subfigure}{.235\textwidth} 
\centering
\includegraphics[width=\textwidth, bb = 10 10 560 450, clip=true]{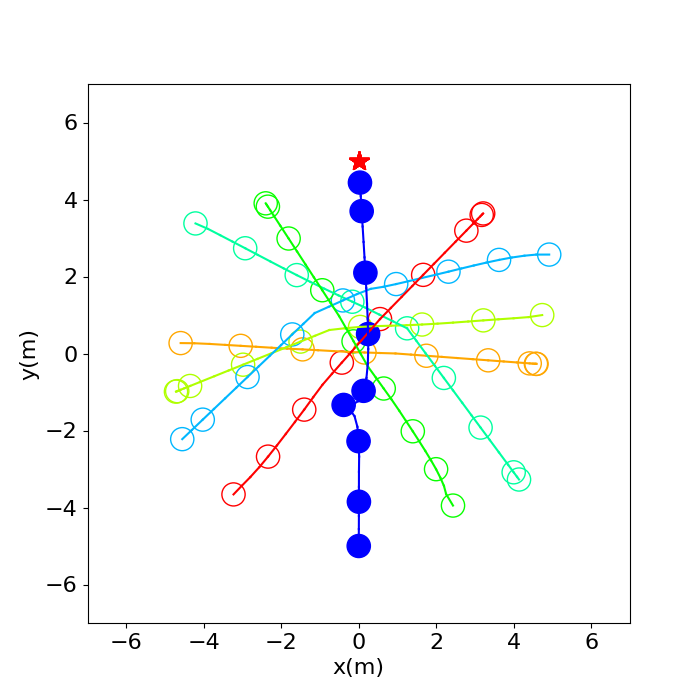}
\caption{$t = \SI{12.4}{s}$}
\end{subfigure} 

\caption{Trajectories of the robot (proposed framework) and human pedestrians at several time-steps in a circle crossing simulation with the robot visible to the humans. The destination of the robot is marked by a red star.}
\label{fig:traj}
\end{figure}

\setlength{\extrarowheight}{.8mm} 
\begin{table*}[h!]
\centering
\caption{Statistical results in Circle and Square Crossing Scenarios - Success  Rates}
 \begin{tabular}{ |c | c || c|c|c|c| c|c|c|c |}
 \hline
  & & \multicolumn{8}{c|}{Success Rate (\%)} \\
 \hline
\multirow{2}{*}{{Method $\downarrow$}{}} & { \# Humans} & 
\multicolumn{2}{c|}{$5$}  & \multicolumn{2}{c|}{$6$} & \multicolumn{2}{c|}{$7$} & \multicolumn{2}{c|}{$8$} \\  
\cline{2-10}
 & {Scenario}& {C $\Circle$  } & {S $\square$  } & {C $\Circle$  } & {S $\square$  }& {C $\Circle$  } & {S $\square$  } & {C $\Circle$  } & {S $\square$  } \\
 \hline \hline
\multicolumn{2}{|c||}{MPC}   &  $\boldsymbol{99.4}$ & $\boldsymbol{99.5}$ & $\boldsymbol{99.6}$ & $\boldsymbol{99.4}$ &
  $\boldsymbol{99}$ & $\boldsymbol{98.7}$ & $\boldsymbol{98.1}$ & $\boldsymbol{98.1}$ \\
 \hline
 \multicolumn{2}{|c||}{CADRL}  & $98.9$ & $32.1$ & $97.6$ & $32.4$  &
$93.8$ & $29.5$ & $94.0$ & $28.9$ \\
 \hline
 \multicolumn{2}{|c||}{SARL}  & $96.8$ & $47.5$ & $97.4$ & $41.3$ 
 & $97.0$ & $35.5$ & $96.7$ & $33.0$ \\

\hline
\end{tabular}
\label{tab1:SuccAndColl}
\vspace{-2mm}
\end{table*}

\begin{table*}[h!]
\centering
\caption{Statistical results in Circle and Square Crossing Scenarios - Collision Rates}
 \begin{tabular}{ |c | c ||  c|c|c|c| c|c|c|c| }
 \hline
  & &  \multicolumn{8}{c|}{Collision Rate (\%)} \\
 \hline
\multirow{2}{*}{{Method $\downarrow$}{}} & { \# Humans} & 
\multicolumn{2}{c|}{$5$}  & \multicolumn{2}{c|}{$6$} & \multicolumn{2}{c|}{$7$} & \multicolumn{2}{c|}{$8$} \\
\cline{2-10}
 & {Scenario}& {C $\Circle$  } & {S $\square$  } & {C $\Circle$  } & {S $\square$  }& {C $\Circle$  } & {S $\square$  } & {C $\Circle$  } & {S $\square$  } \\
 \hline \hline
\multicolumn{2}{|c||}{MPC}   
  &  $0.5$ & $0.2$ & $0.4$ & $0.3$ &
  $0.8$ & $0.6$ & $1.8$ & $0.9$ \\ 
 \hline
 \multicolumn{2}{|c||}{CADRL}  
& $0.1$ & $0.2$ & $\boldsymbol{0.0}$ & $0.3$  &
$\boldsymbol{0.0}$ & $0.3$ & $\boldsymbol{0.0}$ & $0.6$\\
 \hline
 \multicolumn{2}{|c||}{SARL}  
 & $\boldsymbol{0.0}$ & $\boldsymbol{0.0}$ & $0.1$ & $\boldsymbol{0.0}$ 
 & $0.2$ & $\boldsymbol{0.0}$ & $0.2$ & $\boldsymbol{0.0}$\\
\hline
\end{tabular}
\label{tab2:Coll}
\vspace{-2mm}
\end{table*}

\begin{table*}[h!]
\centering
\caption{Statistical results in Circle and Square Crossing Scenarios - Discomfort Rates}
  \begin{tabular}{ |c | c || c|c|c|c| c|c|c|c |}
 \hline
  & & \multicolumn{8}{c|}{Discomfort Rate (\%)} \\
 \hline
 \multirow{2}{*}{{Method $\downarrow$}{}} & { \# Humans} & 
\multicolumn{2}{c|}{$5$}  & \multicolumn{2}{c|}{$6$} & \multicolumn{2}{c|}{$7$} & \multicolumn{2}{c|}{$8$}  \\
\cline{2-10}
 & {Scenario}& {C $\Circle$  } & {S $\square$  } & {C $\Circle$  } & {S $\square$  }& {C $\Circle$  } & {S $\square$  } & {C $\Circle$  } & {S $\square$  } \\
 \hline \hline
\multicolumn{2}{|c||}{MPC}  & $\boldsymbol{0.2}$ &$ \boldsymbol{0.4}$ & $\boldsymbol{0.4}$ & $\boldsymbol{0.4} $
&  $\boldsymbol{0.4}$ & $\boldsymbol{0.7 }$& $\boldsymbol{1.1}$ & $\boldsymbol{0.6}$ 
 \\  \hline
\multicolumn{2}{|c||}{CADRL} & $1.4$ & $4.3$ & $1.6$ &$ 4.7$
&$1.0$ &$ 4.9 $&$ 2.1$ & $6.4$
 \\ \hline
 \multicolumn{2}{|c||}{SARL} & $0.6$ & $1.1$ & $0.8$ & $1.4$ 
 &  $ 2.2$ & $1.1$ &$ 3.1$ & $1.3$
 \\ \hline
\end{tabular}
\label{tab3:Disc}
\end{table*}

\begin{table*}[h!]
\centering
\caption{Statistical results in Circle and Square Crossing Scenarios - Average Travel Time}
  \begin{tabular}{ |c | c || c|c|c|c| c|c|c|c |}
 \hline
  & &  \multicolumn{8}{c|}{Average Travel Time $(s)$} \\
 \hline
 \multirow{2}{*}{{Method $\downarrow$}{}} & { \# Humans} & 
\multicolumn{2}{c|}{$5$}  & \multicolumn{2}{c|}{$6$} & \multicolumn{2}{c|}{$7$} & \multicolumn{2}{c|}{$8$}  \\
\cline{2-10}
 & {Scenario}& {C $\Circle$  } & {S $\square$  } & {C $\Circle$  } & {S $\square$  }& {C $\Circle$  } & {S $\square$  } & {C $\Circle$  } & {S $\square$  } \\
 \hline \hline
\multicolumn{2}{|c||}{MPC}  
& $\boldsymbol{13.4}$ & $\boldsymbol{11.8}$ & $\boldsymbol{14.0}$ & $\boldsymbol{12.2}$ 
&  $14.8$ & $\boldsymbol{12.5}$ & $15.2$ & $\boldsymbol{12.9}$ 
 \\  \hline
\multicolumn{2}{|c||}{CADRL} 
& $14.0$ & $16.0$ & $14.4$ & $16.0$
& $14.9$ & $16.1$ &$ 15.4$ &$ 16.8$
 \\ \hline
 \multicolumn{2}{|c||}{SARL} 
 & $13.8$ & $14.9$ & $\boldsymbol{14.0}$ & $15.3$ 
 &   $\boldsymbol{14.3}$ & $15.8$ & $\boldsymbol{14.7}$ & $16.2$
 \\ \hline
\end{tabular}
\label{tab4:Time}
\end{table*}

The planning algorithm was implemented in Python in which CasADi \cite{andersson2019casadi} and the IPOPT solver \cite{wachter2006implementation} are used for formulating and solving the MPC problem, respectively.

We used the following parameters for the MPC problem
${\tau = \SI{0.4} {s}}$,
${H = 8}$,
${L = 8}$,
${v_\mathrm{max} = \SI{1.0} {m/s}}$,
${a_\mathrm{max} = \SI{2.0} {m/s^2}}$, 
${d_\mathrm{min} = \SI{0.8} {m}}$,
${\rho = \SI{0.5} {s}^2}$,
${\mu = 30}$,
${\omega^\mathrm{goal} = 10.0}$, 
${\omega^\mathrm{acce} = 10^{-1}}$, 
${\omega^\mathrm{jerk} = 10^{-1}}$,
${\omega^\mathrm{coll} = 10^{10}}$.
The simulations were executed on an MSI computer with an Intel Core i9 CPU, \SI{64}{GB} RAM, and a GeForce RTX 3080 Ti GPU. For social navigation simulations, we used the CrowdNav environment\footnotemark \cite{chen2019crowd} in which the human pedestrians are simulated using Optimal Reciprocal Collision Avoidance (ORCA) \cite{van2008reciprocal}. 
We utilize Trajnet++ benchmark\footnotemark \cite{kothari2021human} for training Social-LSTM models using the ETH dataset \cite{pellegrini2009you}. 

\begin{figure}[h!] 
\centering
\begin{subfigure}{.235\textwidth} 
\centering
\includegraphics[width=\textwidth, bb = 10 10 560 450, clip=true]{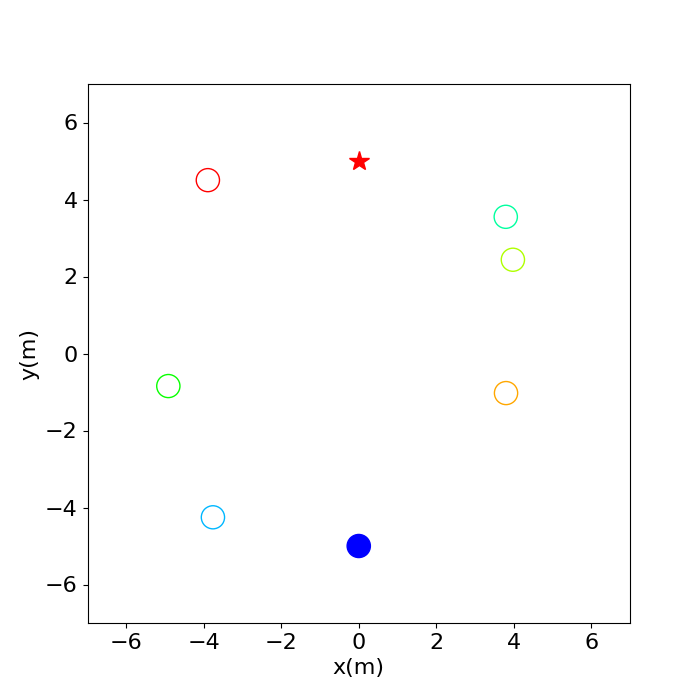}
\caption{$t = \SI{0}{s}$}
\end{subfigure} 
\begin{subfigure}{.23\textwidth} 
\centering
\includegraphics[width=\textwidth, bb = 10 10 560 450, clip=true]{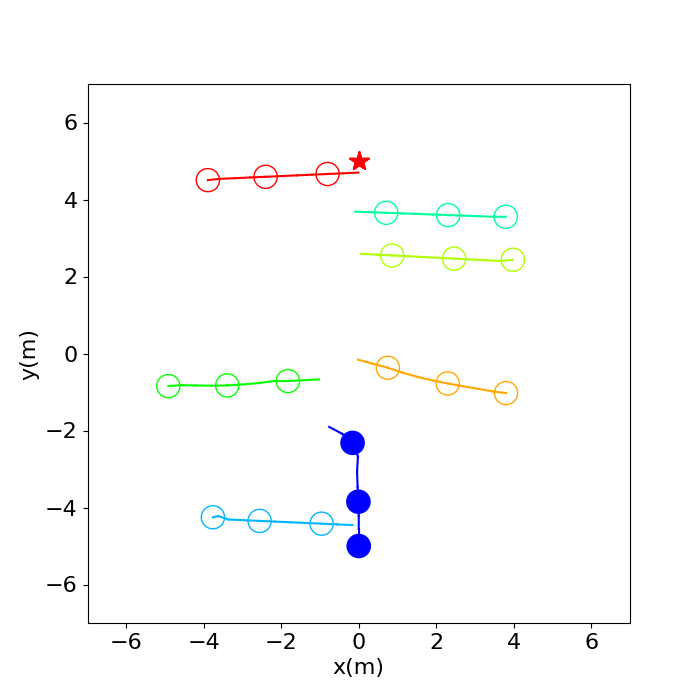}
\caption{$t = \SI{4}{s}$}
\end{subfigure} 

\begin{subfigure}{.235\textwidth} 
\centering
\includegraphics[width=\textwidth, bb = 10 10 560 450, clip=true]{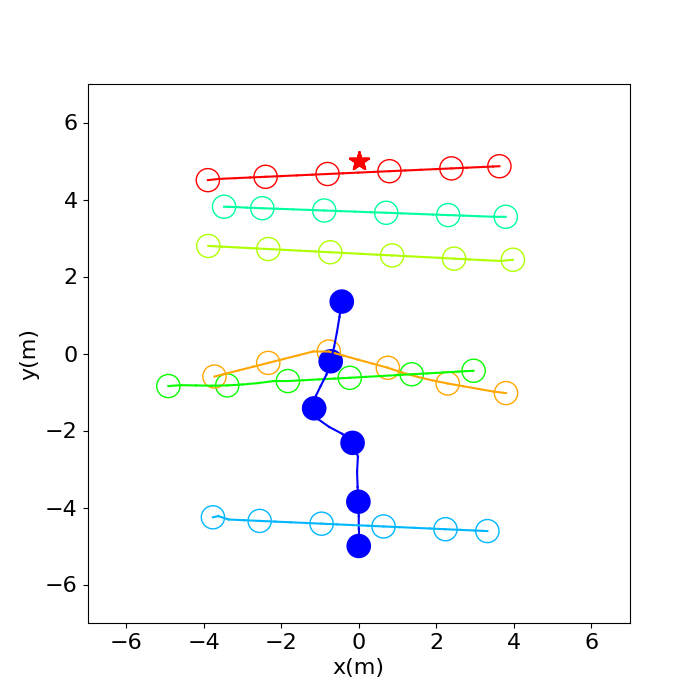}
\caption{$t = \SI{8}{s}$}
\end{subfigure} 
\begin{subfigure}{.235\textwidth} 
\centering
\includegraphics[width=\textwidth, bb = 10 10 560 450, clip=true]{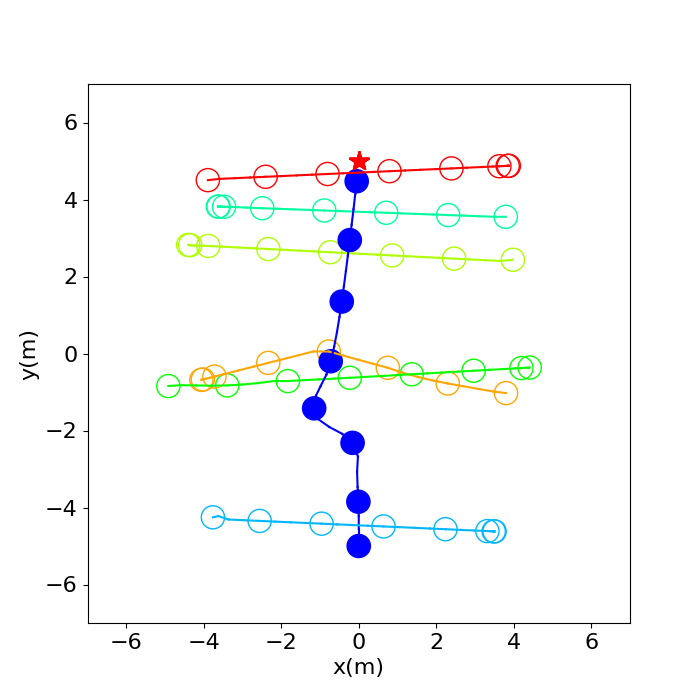}
\caption{$t = \SI{11.6}{s}$}
\end{subfigure} 

\caption{Trajectories of the robot (proposed framework) and human pedestrians at several time-steps in a square crossing simulation with the robot visible to the humans. The destination of the robot is marked by a red star.}
\label{fig:traj_square}
\end{figure}

We demonstrate the effectiveness of the proposed method by the trajectories of the robot and human pedestrians in a circle crossing simulation with $6$ human pedestrians in Fig.~\ref{fig:traj}.
As shown in the figure, the robot can effectively navigate among the humans and reach the goal in $\SI{15.2}{s}$ without colliding with any of them. 
{The robot is visible to the humans during motion to ensure that the interactive behaviors are captured by the prediction module. Similarly, the trajectories of a single robot as it navigates among $6$ pedestrians in a square crossing scenario is depicted in Fig. \ref{fig:traj_square}. Notably, the robot is able to reach its destination successfully at  $\SI{11.6}{s}$ without violating any safety constraints.}

\footnotetext[1]{\url{https://github.com/vita-epfl/CrowdNav}}

\footnotetext[2]{\url{https://github.com/vita-epfl/trajnetplusplusbaselines}}

To further validate the performance of the proposed method in comparison with different navigation algorithms,
we collect and compare the following metrics: 
\begin{itemize}
\item \textbf{Success rate}: {The percentage of simulations in which the robot successfully reaches its respective destinations.}
\item \textbf{Collision rate}: {The percentage of simulations in which the minimum distance between the robot and the pedestrians is below \SI{0.8}{m}, indicating a violation of personal space.}
\item \textbf{Discomfort rate}: The percentage of simulations that the robot’s projected path intersects with a pedestrian’s projected path \cite{wang2022metrics}. 
The projected path is defined as a line segment from the current position along with the direction of the velocity and the length proportional to the speed.
\item \textbf{Average travel time}: {The time it takes for the robot to reach its destination averaged over all successful simulations.}
\end{itemize}

Among the four metrics, the success rate and average time to the destination describe the path quality of the navigation algorithms. 
On the other hand, the collision and discomfort rates are related to social conformity \cite{wang2022metrics}. 
If the simulation reports neither a success nor a collision,  it means a timeout has occurred wherein the robot has not been able to traverse to its destination. 

We compare our proposed MPC with two different RL algorithms including CADRL \cite{chen2017socially} and SARL \cite{chen2019crowd}. 
This evaluation was based on $8000$ simulations with varying numbers of human agents and randomized initial conditions. 
We trained both RL algorithms using the circle-crossing scenario in the CrowdNav environment with a similar implementation and parameters of \cite{chen2019crowd} including the imitation learning step. {However, to thoroughly evaluate their generalization capability, we conducted tests in both circle-crossing and square-crossing scenarios, and summarized our results in Tables \ref{tab1:SuccAndColl}-\ref{tab4:Time}.}

{Table \ref{tab1:SuccAndColl} depicts the success rate of a robot navigating in both circle and square crossing scenarios among pedestrians, crowd sizes ranging from $5$ to $8$. 
We conducted $1000$ simulations for each combination of scenarios and number of human pedestrians, using randomized initial conditions generated from random seeds that were different from the training seeds for CADRL and SARL. 
As can be seen, our proposed approach consistently yielded a greater success rate in comparison to the RL-based strategies across all scenarios. While the performance of CADRL and SARL is comparable to our proposed approach in the circle crossing scenario, their success rate drops over $50$\% in the square crossing scenario which is different from their training environment. The decrease in success rate is due to a significant timeout rate, indicating the incapability of the policies to generate a feasible path. 
This clearly illustrates the significant reliance of RL techniques on the domains in which they have been trained. 
}

{Tables \ref{tab2:Coll} and \ref{tab3:Disc} summarize the collision and discomfort rates for our proposed approach and RL-based navigation methods. Based on Table \ref{tab2:Coll}, it can be observed that MPC results in collision rates of less than $2\%$ across all scenarios. 
For RL-based navigation methods, the collision rate is less than $0.6\%$, however, they exhibit a higher discomfort rate than the MPC approach as shown in Table \ref{tab3:Disc}. In particular, the CADRL policy encounters growing discomfort rates with increasing crowd densities. 
MPC on the other hand, shows a lower discomfort rate, which implies MPC approach can {be more socially conscious} of the human pedestrians' motion. }

In Table \ref{tab4:Time}, we provide results of average travel time for the MPC approach and RL-based methods. 
In almost all scenarios (except two), we observe higher travel times for the RL techniques. 
Overall, the performance of MPC and RL algorithms in circle crossing simulations is highly comparable. 
Unlike the RL algorithms, the control policy in the proposed MPC formulation does not need any pre-training. 
Furthermore, RL algorithms are highly susceptible to domain shift. 
{As mentioned earlier, for this set of evaluations,} RL-based methods have been trained in the circle-crossing scenario but tested in both the circle-crossing and a square-crossing scenario.

{Next, we investigated the sensitivities of the metrics to the different parameters used in the objective function for the proposed approach. 
We ran four sets of $100$ simulations with varying crowd sizes of $5$ and $10$ pedestrians each in a circle crossing scenario. 
The parameter $w^{coll}$ representing the weight on collision avoidance, denoted as $w^{coll}$, was held constant at a value of $10^{10}$.
When $w^{goal}$ is set to a low value of $0.1$, we observe a significant increase in average travel time on the introduction of penalties on acceleration and jerk. 
However, this effect is nullified on increasing $w^{goal}$ to $10$.
This observation is consistent across different crowd sizes. 
Moreover, as the crowd density increases, there is a marked increase in the travel time and a slight decrease in the success rates. 
The difficulty induced due to the multi-objective nature of the optimization is apparent in the way these costs compete against one another in high crowd densities. }

\begin{table}[h!]
\caption{The effects of parameters on crowd navigation performance in a circle crossing scenario over $100$ simulations }
\centering
 \begin{tabular}{|c|c|c| c|c|c|c|}
 \hline
\#&$w^{acce}$  & $w^{goal}$ &Suc. & Coll.&Disc.&Avg. Time\\
Humans& $w^{jerk}$ & & (\%)& (\%)&(\%)& (s)\\
\hline
5&$0$& $0.1$  &$100$ & $0$&$0$ &$13.5$\\ \hline
5&$0.1$& $0.1$ &$100$ & $0$&$0$ &$18.7$\\ \hline
5&$0$& $10$ &$100$ & $0$&$1$ &$13.2$\\ \hline
5&$0.1$ & $10$  &$100$ &$0$&$0$&$13.3$\\\hline
\hline
10&$0$ & $0.1$  &$99$ &$0$&$0$&$16.6$\\\hline
10&$0.1$ & $0.1$  &$96$ &$0$&$0$&$20.3$\\\hline
10&$0$ & $10$ & $99$& $0$&$0$ &$16.3$  \\\hline
10&$0.1$ & $10$  &$98$ &$2$&$1$&$16.1$\\\hline
\end{tabular}
\label{tab1:SA}
\end{table}

{To assess the effects of horizon length on the performance, we conducted another set of experiments for circle and square crossing scenarios. 
For each combination of scenario and control horizon length, we performed $100$ simulations with a crowd size of $6$ and present the findings in Table \ref{tab:compTimeVsH}. 
Consistent with our expectations, our finding indicates that a longer control horizon has a direct positive impact on the success rate. However, it is evident that there exists a trade-off between the extension of the control horizon and the cost of computational processes. 
In this case, $8$ turns out to be a suitable choice for the control horizon. }

\begin{table}[h!]
    \centering
        \caption{Computation Time Increase with Control Horizon }
\begin{tabular}{|c|c |c| c| c| c|}
    \hline
     Scenario & H & Comp. Time (s) & Suc. & Coll. & Disc. \\
     \hline
     {C $\Circle$  }& $4$ & $0.11$ & $97$ & $2$ & $2$\\
     \hline
     {S $\square$  } & $4$& $0.1$ & $96$ & $0$ & $1$\\
     \hline
     {C $\Circle$  }& $8$ & $0.26$ & $100$ & $0$ & $0$\\
     \hline
     {S $\square$  } & $8$& $0.22$ & $99$ & $0$ & $1$\\
     \hline
     {C $\Circle$  }& $12$ & $0.56$ & $100$ & $0$ & $1$\\
     \hline
     {S $\square$  } & $12$ & $0.47$ & $100$ & $0$ & $1$\\
     \hline
\end{tabular}
    \label{tab:compTimeVsH}
\end{table}

\section{Conclusions}
\label{sec:conclu}
This work presented a control framework for navigating an individual robot in crowded environments.
Our control framework is a combination of MPC and a human trajectory prediction model based on Social-LSTM.
In order to evaluate the performance of our proposed approach, we conducted extensive simulations and compared our approach against several state-of-the-art RL algorithms. 
We showed that the performance of our proposed approach in benchmark scenarios are highly comparable, and in contrast to the RL algorithms, our control policy is not liable to suffer from a distribution shift. 
{A possible extension is to investigate the effectiveness of the proposed control framework in dealing with multiple robots navigating in a coordinated manner through crowds of humans. }
Additionally, to bridge the gap between simulation and reality, another future research direction is to investigate the variety of human behaviors and to incorporate more measures of sociability into the planning framework.

\bibliographystyle{IEEEtran.bst} 
\bibliography{reference/ref.bib}

\begin{thebibliography}{10}
\providecommand{\url}[1]{#1}
\csname url@rmstyle\endcsname
\providecommand{\newblock}{\relax}
\providecommand{\bibinfo}[2]{#2}
\providecommand\BIBentrySTDinterwordspacing{\spaceskip=0pt\relax}
\providecommand\BIBentryALTinterwordstretchfactor{4}
\providecommand\BIBentryALTinterwordspacing{\spaceskip=\fontdimen2\font plus
\BIBentryALTinterwordstretchfactor\fontdimen3\font minus \fontdimen4\font\relax}
\providecommand\BIBforeignlanguage[2]{{%
\expandafter\ifx\csname l@#1\endcsname\relax
\typeout{** WARNING: IEEEtran.bst: No hyphenation pattern has been}%
\typeout{** loaded for the language `#1'. Using the pattern for}%
\typeout{** the default language instead.}%
\else
\language=\csname l@#1\endcsname
\fi
#2}}

\bibitem{KRUSE20131726}
T.~Kruse, A.~K. Pandey, R.~Alami, and A.~Kirsch, ``Human-aware robot navigation: A survey,'' \emph{Robotics and Autonomous Systems}, vol.~61, no.~12, pp. 1726--1743, 2013.

\bibitem{CHARALAMPOUS201785}
K.~Charalampous, I.~Kostavelis, and A.~Gasteratos, ``Recent trends in social aware robot navigation: A survey,'' \emph{Robotics and Autonomous Systems}, vol.~93, pp. 85--104, 2017.

\bibitem{francis2023principles}
A.~Francis, C.~Pérez-D'Arpino, C.~Li, F.~Xia, A.~Alahi, R.~Alami, A.~Bera, A.~Biswas, J.~Biswas, R.~Chandra, H.-T.~L. Chiang, M.~Everett, S.~Ha, J.~Hart, J.~P. How, H.~Karnan, T.-W.~E. Lee, L.~J. Manso, R.~Mirksy, S.~Pirk, P.~T. Singamaneni, P.~Stone, A.~V. Taylor, P.~Trautman, N.~Tsoi, M.~Vázquez, X.~Xiao, P.~Xu, N.~Yokoyama, A.~Toshev, and R.~Martín-Martín, ``Principles and guidelines for evaluating social robot navigation algorithms,'' 2023.

\bibitem{mavrogiannis2023core}
C.~Mavrogiannis, F.~Baldini, A.~Wang, D.~Zhao, P.~Trautman, A.~Steinfeld, and J.~Oh, ``Core challenges of social robot navigation: A survey,'' \emph{ACM Transactions on Human-Robot Interaction}, vol.~12, no.~3, pp. 1--39, 2023.

\bibitem{van2008reciprocal}
J.~Van~den Berg, M.~Lin, and D.~Manocha, ``Reciprocal velocity obstacles for real-time multi-agent navigation,'' in \emph{2008 IEEE international conference on robotics and automation}.\hskip 1em plus 0.5em minus 0.4em\relax Ieee, 2008, pp. 1928--1935.

\bibitem{chen2017socially}
Y.~F. Chen, M.~Everett, M.~Liu, and J.~P. How, ``Socially aware motion planning with deep reinforcement learning,'' in \emph{2017 IEEE/RSJ International Conference on Intelligent Robots and Systems (IROS)}.\hskip 1em plus 0.5em minus 0.4em\relax IEEE, 2017, pp. 1343--1350.

\bibitem{everett2018motion}
M.~Everett, Y.~F. Chen, and J.~P. How, ``Motion planning among dynamic, decision-making agents with deep reinforcement learning,'' in \emph{2018 IEEE/RSJ International Conference on Intelligent Robots and Systems (IROS)}.\hskip 1em plus 0.5em minus 0.4em\relax IEEE, 2018, pp. 3052--3059.

\bibitem{chen2019crowd}
C.~Chen, Y.~Liu, S.~Kreiss, and A.~Alahi, ``Crowd-robot interaction: Crowd-aware robot navigation with attention-based deep reinforcement learning,'' in \emph{2019 international conference on robotics and automation (ICRA)}.\hskip 1em plus 0.5em minus 0.4em\relax IEEE, 2019, pp. 6015--6022.

\bibitem{liu2023intention}
S.~Liu, P.~Chang, Z.~Huang, N.~Chakraborty, K.~Hong, W.~Liang, D.~L. McPherson, J.~Geng, and K.~Driggs-Campbell, ``Intention aware robot crowd navigation with attention-based interaction graph,'' in \emph{2023 IEEE International Conference on Robotics and Automation (ICRA)}.\hskip 1em plus 0.5em minus 0.4em\relax IEEE, 2023, pp. 12\,015--12\,021.

\bibitem{brito2021go}
B.~Brito, M.~Everett, J.~P. How, and J.~Alonso-Mora, ``Where to go next: Learning a subgoal recommendation policy for navigation in dynamic environments,'' \emph{IEEE Robotics and Automation Letters}, vol.~6, no.~3, pp. 4616--4623, 2021.

\bibitem{akhtyamov2023social}
T.~Akhtyamov, A.~Kashirin, A.~Postnikov, and G.~Ferrer, ``Social robot navigation through constrained optimization: a comparative study of uncertainty-based objectives and constraints,'' \emph{arXiv preprint arXiv:2305.02859}, 2023.

\bibitem{chen2021interactive}
Y.~Chen, F.~Zhao, and Y.~Lou, ``Interactive model predictive control for robot navigation in dense crowds,'' \emph{IEEE Transactions on Systems, Man, and Cybernetics: Systems}, vol.~52, no.~4, pp. 2289--2301, 2021.

\bibitem{poddar2023crowd}
S.~Poddar, C.~Mavrogiannis, and S.~S. Srinivasa, ``From crowd motion prediction to robot navigation in crowds,'' \emph{arXiv preprint arXiv:2303.01424}, 2023.

\bibitem{lindemann2022safe}
L.~Lindemann, M.~Cleaveland, G.~Shim, and G.~J. Pappas, ``Safe planning in dynamic environments using conformal prediction,'' \emph{IEEE Robotics and Automation Letters}, 2023.

\bibitem{vulcano2022safe}
V.~Vulcano, S.~G. Tarantos, P.~Ferrari, and G.~Oriolo, ``Safe robot navigation in a crowd combining nmpc and control barrier functions,'' in \emph{2022 IEEE 61st Conference on Decision and Control (CDC)}.\hskip 1em plus 0.5em minus 0.4em\relax IEEE, 2022, pp. 3321--3328.

\bibitem{alahi2016social}
A.~Alahi, K.~Goel, V.~Ramanathan, A.~Robicquet, L.~Fei-Fei, and S.~Savarese, ``Social lstm: Human trajectory prediction in crowded spaces,'' in \emph{Proceedings of the IEEE conference on computer vision and pattern recognition}, 2016, pp. 961--971.

\bibitem{gupta2018social}
A.~Gupta, J.~Johnson, L.~Fei-Fei, S.~Savarese, and A.~Alahi, ``Social gan: Socially acceptable trajectories with generative adversarial networks,'' in \emph{Proceedings of the IEEE conference on computer vision and pattern recognition}, 2018, pp. 2255--2264.

\bibitem{liu2021social}
Y.~Liu, Q.~Yan, and A.~Alahi, ``Social nce: Contrastive learning of socially-aware motion representations,'' in \emph{Proceedings of the IEEE/CVF International Conference on Computer Vision}, 2021, pp. 15\,118--15\,129.

\bibitem{trautman2017sparse}
P.~Trautman, ``Sparse interacting gaussian processes: Efficiency and optimality theorems of autonomous crowd navigation,'' in \emph{2017 IEEE 56th Annual Conference on Decision and Control (CDC)}.\hskip 1em plus 0.5em minus 0.4em\relax IEEE, 2017, pp. 327--334.

\bibitem{trautman2015robot}
P.~Trautman, J.~Ma, R.~M. Murray, and A.~Krause, ``Robot navigation in dense human crowds: Statistical models and experimental studies of human--robot cooperation,'' \emph{The International Journal of Robotics Research}, vol.~34, no.~3, pp. 335--356, 2015.

\bibitem{espinoza2022deep}
J.~L.~V. Espinoza, A.~Liniger, W.~Schwarting, D.~Rus, and L.~Van~Gool, ``Deep interactive motion prediction and planning: Playing games with motion prediction models,'' in \emph{Learning for Dynamics and Control Conference}.\hskip 1em plus 0.5em minus 0.4em\relax PMLR, 2022, pp. 1006--1019.

\bibitem{bacsar1998dynamic}
T.~Ba{\c{s}}ar and G.~J. Olsder, \emph{Dynamic noncooperative game theory}.\hskip 1em plus 0.5em minus 0.4em\relax SIAM, 1998.

\bibitem{wang2022metrics}
J.~Wang, W.~P. Chan, P.~Carreno-Medrano, A.~Cosgun, and E.~Croft, ``Metrics for evaluating social conformity of crowd navigation algorithms,'' in \emph{2022 IEEE International Conference on Advanced Robotics and Its Social Impacts (ARSO)}.\hskip 1em plus 0.5em minus 0.4em\relax IEEE, 2022, pp. 1--6.

\bibitem{gupta2023interaction}
P.~Gupta, D.~Isele, D.~Lee, and S.~Bae, ``Interaction-aware trajectory planning for autonomous vehicles with analytic integration of neural networks into model predictive control,'' \emph{arXiv preprint arXiv:2301.05393}, 2023.

\bibitem{kothari2021human}
P.~Kothari, S.~Kreiss, and A.~Alahi, ``Human trajectory forecasting in crowds: A deep learning perspective,'' \emph{IEEE Transactions on Intelligent Transportation Systems}, vol.~23, no.~7, pp. 7386--7400, 2021.

\bibitem{williams2018best}
G.~Williams, B.~Goldfain, P.~Drews, J.~M. Rehg, and E.~A. Theodorou, ``Best response model predictive control for agile interactions between autonomous ground vehicles,'' in \emph{2018 IEEE International Conference on Robotics and Automation (ICRA)}.\hskip 1em plus 0.5em minus 0.4em\relax IEEE, 2018, pp. 2403--2410.

\bibitem{andersson2019casadi}
J.~A. Andersson, J.~Gillis, G.~Horn, J.~B. Rawlings, and M.~Diehl, ``Casadi: a software framework for nonlinear optimization and optimal control,'' \emph{Mathematical Programming Computation}, vol.~11, pp. 1--36, 2019.

\bibitem{wachter2006implementation}
A.~W{\"a}chter and L.~T. Biegler, ``On the implementation of an interior-point filter line-search algorithm for large-scale nonlinear programming,'' \emph{Mathematical programming}, vol. 106, pp. 25--57, 2006.

\bibitem{pellegrini2009you}
S.~Pellegrini, A.~Ess, K.~Schindler, and L.~Van~Gool, ``You'll never walk alone: Modeling social behavior for multi-target tracking,'' in \emph{2009 IEEE 12th international conference on computer vision}.\hskip 1em plus 0.5em minus 0.4em\relax IEEE, 2009, pp. 261--268.

\end{thebibliography}


\end{document}